# Diagnostic Communication and Visual System based on Vehicle UDS Protocol

Hong Zhang，Ding Li

Harbin Institute of Technology

*Abstract*—Unified Diagnostic Services (UDS) is a diagnostic communication protocol used in electronic control units (ECUs) within automotive electronics, which is specified in the ISO 14229-1. It is derived from ISO 14230-3 (KWP2000) and the now obsolete ISO 15765-3 (Diagnostic Communication over Controller Area Network (DoCAN). 'Unified' in this context means that it is an international and not a company-specific standard. By now this communication protocol is used in all new ECUs made by Tier 1 suppliers of Original Equipment Manufacturer (OEM), and is incorporated into other standards, such as AUTOSAR. The ECUs in modern vehicles control nearly all functions, including electronic fuel injection (EFI), engine control, the transmission, anti-lock braking system, door locks, braking, window operation, and more.

*Index terms*—UDS，Diagnostic，Bootloader，Data

■ THE INTRODUCTION

As evident, UDS enables extensive control over the vehicle ECUs. For security reasons, critical UDS services are therefore restricted through an authentication process. Basically, an ECU will send a 'seed' to a tester, who in turn must produce a 'key' to gain access to security-critical services. To retain this access, the tester needs to send a 'tester present' message periodically. In practice, this authentication process enables vehicle manufactures to restrict UDS access for aftermarket users and ensure that only designated tools will be able to utilize the security-critical UDS services. Note that the switching between authentication levels is done through diagnostic session control, which is one of the UDS services available. Vehicle manufactures can decide which sessions are supported, though they must always support the 'default session' (i.e. which does not involve any authentication). With that said, they decide what services are supported within the default session as well. If a tester tool switches to a non-default session, it must send a 'tester present' message periodically to avoid being returned to the default session. Data flow [1][9] in communication process is then shared with other ECUs.

Department Head

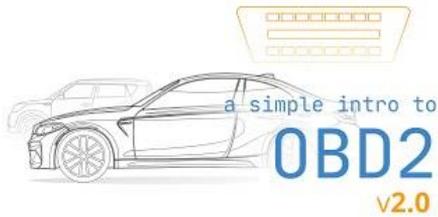

■ BACKGROUND

Generally, when a request is sent to an ECU [2][3], the ECU may respond positively or negatively. In case the response is positive, the tester may want to suppress the response (as it may be irrelevant). This is done by setting the 1st bit to 1 in the sub function byte. Negative responses cannot be suppressed.

The Controller Area Network - CAN bus [4][8] is a message-based protocol designed to allow the Electronic Control Units (ECUs) found in today's automobiles, as well as other devices, to communicate with each other in a reliable, priority-driven fashion. Messages or "frames" are received by all devices in the network, which does not require a host computer. CAN is supported by a rich set of international standards under ISO 11898 [17][20].

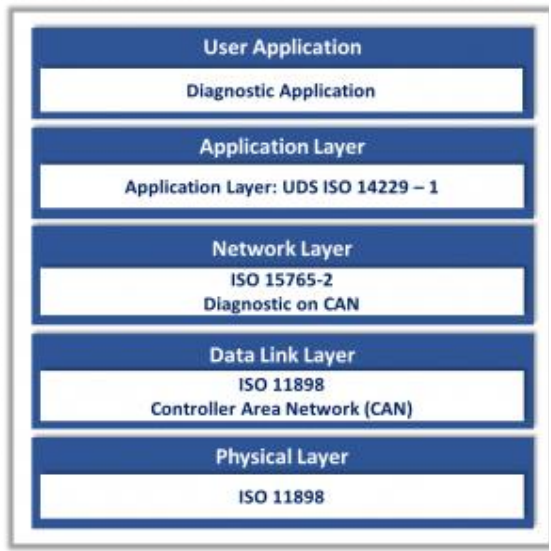

Vehicle diagnostics as a process has undergone numerous transformations over the past 2 decades. The demand for a more accurate, standard, and efficient fault detection in vehicle diagnostics, has led to breakthrough innovations and developments.

If you're looking to request UDS-based diagnostic trouble codes (DTC) [5][16], you'll of course have to use UDS communication for this purpose. However, if your aim is to record current sensor values it is less clear. Typically, data parameters of interest for e.g. vehicle telematics (speed, state of charge etc) will already be communicated between ECUs on the CAN bus in the form of raw CAN frames - without the need for a diagnostic tool requesting this information. That is because ECUs rely on communicating this information as part of their operation (as explained in our intro to CAN bus). If you have direct access to the CAN bus, it would thus appear easier to simply log this raw CAN traffic and process it. If you are the vehicle manufacturer, you will know how to interpret this raw CAN data either way - and it'll be simpler to perform your device configuration and post processing in most cases. If you're in the aftermarket, it'll also be simpler to reverse engineer the raw CAN frames as you can focus on single frames - and avoid the request/response layer of complexity. However, one key reason why UDS is frequently used for extracting sensor values despite the above is due to 'gateways'. Specifically, an increasing share of modern cars have started to block the access to the raw CAN bus data via the OBD2 connector. This is particularly often the case for German vehicles, as well as electric vehicles. To record the existing CAN traffic in such a car, you would need to e.g. use a CAN Crocodile adapter and 'snap' it onto the CAN low/high wiring harness. This in turn will require exposing the wiring harness by removing a panel, which is often prohibitive for many use cases. In contrast, the OBD2 connector gateways typically still allow for UDS communication - incl. sensor value communication. A secondary - and more subtle - reason is that most reverse engineering work is done by 'OBD2 dongle manufacturers' [6][13]. These develop tools that let you extract data across many different cars through the OBD2 connector. Increasingly, the only way for these dongles to get useful information through the OBD2 connector is through UDS communication, which drives a proportionally higher availability of information/databases related to UDS parameters vs. raw CAN parameters.

With the help of the read DTC Information service in the UDS protocol, the testing device can not only read diagnostic related DTC data, but it can also read additional parameters of the component at the time of the occurrence of fault. This helps in pinpointing the root cause of fault/damage and then undertaking the right repair and maintenance operations.

■ TECHNICAL DIRECTION

Record time-synchronous data from ECUs, ADAS sensors [7][15] (video, radar, lidar, ..), bus systems,

analog measurement variables, and much more Unlimited measurement files thanks to the ASAM measurement data format MDF 4.x Maximize data transfer from ECUs through optimized DAQ lists Analyze bus communication in the trace window Calculate additional quantities during the measurement, such as the electrical power [10][18] in the inverter Minimize the amount of data through sophisticated trigger options with pre- and post-trigger times Various window types and user-definable panels are available for graphical display.

OBD2 Dongles: While not a "formal" definition, OBD dongles typically refer to small, low cost and simple-to-use consumer-oriented Bluetooth OBD2 readers. They typically provide data via e.g. a smartphone app, allowing you to get a real-time view of your vehicle's performance [11][19]. They are great for plug-and-play consumer purposes but offer very limited flexibility in terms of use cases. Typically, these devices use an ELM327 microcontroller.

MOST is used in nearly every car brand around the world. Up to 64 devices can be connected to a MOST ring network, which allows devices to be connected or disconnected easily. Other topologies are also possible, including virtual stars.

The message identifier can be 11-bit (Standard CAN, 2048 different message identifiers) or 29 bit in length (Extended CAN, 537 million different message identifiers). The remote transmission request bit is dominant and indicates that data is being transmitted.

Alteration of a message length byte [12][14], sub-function byte without and without SPR-bit set, and data length code (DLC) has been performed. Alteration for services without sub-function also have been performed. This process is done for services in both default and non-default sessions. This is because some services are implemented in default sessions, while some are implemented in non-default sessions.

However, when dual pairs of parallel data lines are used, this provides redundancy: when a line is damaged, the second line can take over. This is important in mission-critical applications like steering and braking. FlexRay applications that are not mission-critical typically use a single twisted pair.

According to Gartner, in 2017 there were a total of 19.3 million ethernet ports installed in consumer vehicles. By 2020 this has risen to 122.8 million, a number that is projected to double by 2023.

New technologies such as driver assistance and even self-driving/autonomous vehicle functionalities require higher and higher bandwidth in order to work. This need for speed, coupled with the low cost of Ethernet hardware, has been a big factor in promoting Automotive Ethernet among carmakers. Other motivations for automotive ethernet include the transfer rates needed for LIDAR and other sensors, raw camera data, GPS data, map data, and higher and higher resolution flatscreen displays.